\documentclass{article}
\usepackage{spconf,amsmath,graphicx}
\usepackage{booktabs,spconf}
\usepackage{color,multirow}
\usepackage{amssymb}

\title{Graph Pattern Loss based Diversified Attention Network for Cross-Modal Retrieval}
\name{Xueying~Chen$^1$, Rong~Zhang$^1$, Yibing~Zhan$^2$}
\address{$^1$Department of Electronic Engineering and Information Science,\\
University of Science and Technology of China, Hefei,  China\\
$^2$Hangzhou Dianzi University
}
\begin{document}
\ninept
\maketitle
\begin{abstract}
Cross-modal retrieval aims to enable flexible retrieval experience by combining multimedia data such as image, video, text, and audio. One core of unsupervised approaches is to dig the correlations among different object representations to complete satisfied retrieval performance without requiring expensive labels. In this paper, we propose a Graph Pattern Loss based Diversified Attention Network (GPLDAN) for unsupervised cross-modal retrieval to deeply analyze correlations among representations. First, we propose a diversified attention feature projector by considering the interaction between different representations to generate multiple representations of an instance. Then, we design a novel graph pattern loss to explore the correlations among different representations, in this graph all possible distances between different representations are considered. In addition, a modality classifier is added to explicitly declare the corresponding modalities of features before fusion and guide the network to enhance discrimination ability. We test GPLDAN on four public datasets. Compared with the state-of-the-art cross-modal retrieval methods, the experimental results demonstrate the performance and competitiveness of GPLDAN.
\end{abstract}
\begin{keywords}
Cross-modal retrieval, unsupervised, graph pattern poss, diversified attention
\end{keywords}
\section{Introduction}
\label{sec:intro}

Cross-modal retrieval takes one type of data as the query to retrieve another type of related data \cite{wang2017adversarial}. The core of cross-modal retrieval is to find the relationship between different representations by learning a common representation subspace. With the rapid growth of multimedia data including image, text, and video on the Internet, cross-modal retrieval plays an increasingly important role in many applications \cite{wang2016comprehensive}. However, different modalities usually have different representations and distributions, which is the so-called heterogeneity and makes the similarity of cross-modal representations difficult to calculate. Therefore, the challenge of cross-modal retrieval is how to efficiently calculate the similarity of different modal representations.

To solve the problems mentioned above, several methods have been proposed for cross-modal retrieval in the past several years. Generally, these methods can be divided into three categories according to the need for label information: the supervised methods \cite{wang2017adversarial, peng2017ccl, rasiwasia2010new, wang2015joint, wang2016learning, zhai2012cross, Zhen_2019_CVPR}, the semi-supervised methods \cite{peng2015semi, zhai2013learning}, and the unsupervised methods \cite{hotelling1992relations, andrew2013deep, feng2014cross, he2017unsupervised, wang2015deep, wang2014effective, yan2015deep, zhan2018comprehensive}. Both the supervised methods and the semi-supervised methods use semantic category labels and achieve some results, such as Cross-Modality Correlation Propagation (CMCP) \cite{zhai2012cross}, Adversarial Cross-Modal Retrieval (ACMR) \cite{wang2017adversarial}, Deep Supervised Cross-modal Retrieval (DSCMR) \cite{Zhen_2019_CVPR} and Joint Representation Learning (JRL) \cite{zhai2013learning}. Usually, the label information is difficult to obtain in practical applications, the unsupervised method can alleviate this problem. Among the unsupervised methods, Canonical Correlation Analysis (CCA) \cite{hotelling1992relations} and its variations (e.g., Deep Canonical Correlation Analysis (DCCA) \cite{andrew2013deep} and Deep Canonically Correlated Autoencoders (DCCAE) \cite{wang2015deep}) are ones of the most representative works. Autoencoder based unsupervised methods, such as the Correspondence Autoencoders (Corr-AE) \cite{feng2014cross}, Comprehensive Distance-Preserving Autoencoders (CDPAE) \cite{zhan2018comprehensive} that considers heterogeneous distances and homogeneous distances of representations. In \cite{he2017unsupervised}, the authors proposed UCLA to generate modality-invariant transforms. However despite many improvements that have been made in previous unsupervised approaches and some methods consider some related distances, there still exist the following problems: first, how to effectively use the interaction between different features; second, how to define a comprehensive and effective measurement for the correlations among different representations.

In this paper, we propose a Graph Pattern Loss based Diversified Attention Network (GPLDAN) for unsupervised cross-modal retrieval. To effectively use the interaction between different features and extract better representations, we compute multiple representations of an instance by diversified attention including self-attention and co-attention. To explicitly calculate all possible distances and explore the relationships between these distances, we propose a novel graph pattern loss, in which all the relationships between different representations are considered, regardless of which modality they come from and whether they belong to the same object. The graph pattern loss includes pairwise distance loss, unpaired distance preserving loss and mutual distance preserving loss. The pairwise distance represents the similarity between the two representations of the same object from cross-modalities. The unpaired distance reflects the difference between the two representations of different objects from cross-modalities or the same modalities. The mutual distance preserving loss is to further emphasize the relationship between cross-modal instances. To further help the learning of cross-modal, a modality classifier is added to explicitly declare the corresponding modalities of features before fusion, which is built based on the adversarial learning \cite{goodfellow2014generative}. Compared with other state-of-the-art methods, the experimental results reveal the performance and competitiveness of GPLDAN.

\begin{figure*}[tb]
\setlength{\abovecaptionskip}{0.cm}
\setlength{\belowcaptionskip}{-0.cm}
     \begin{minipage}{1\linewidth}
            \centering 
	    \includegraphics[width=6.5in]{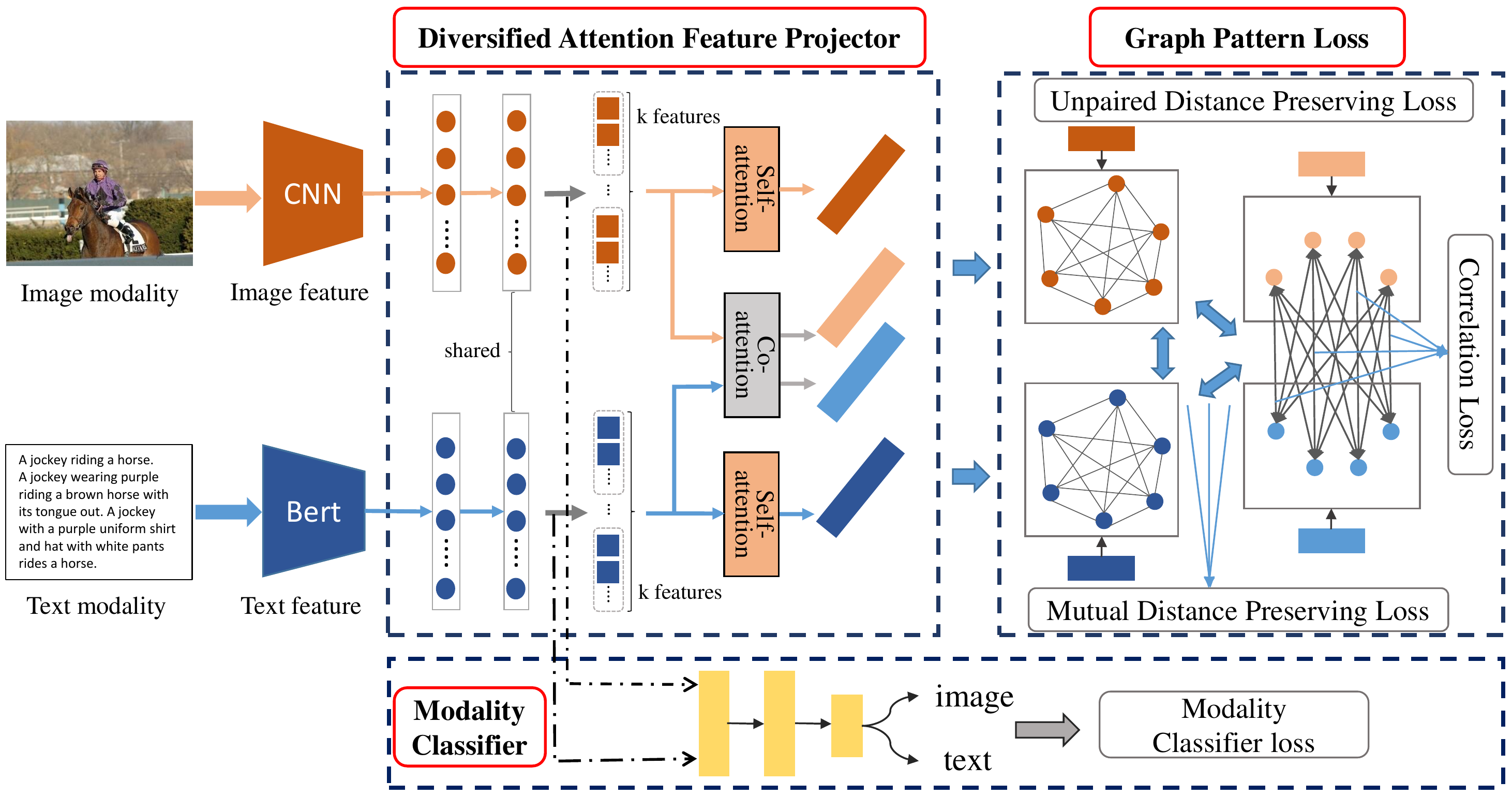} 
     \end{minipage}
\caption{The overall architecture of the proposed GPLDAN. \textbf{Diversified Attention Feature Projector} is to extract multiple representations of an instance by self-attention and co-attention. \textbf{Graph Pattern Loss} includes pairwise distance loss, unpaired distance preserving loss and mutual distance preserving loss. \textbf{Modality Classifier} is to explicitly declare the corresponding modalities of features before fusion.}
\label{fig:network}
\end{figure*}

\section{Method}
\label{sec:method}

The Graph Pattern Loss based Diversified Attention Network (GPLDAN) consists of three parts. Diversified attention feature projector extracts multiple representations of an instance with a diversified attention mechanism. Graph pattern loss includes pairwise distance loss, unpaired distance preserving loss and mutual distance preserving loss, which considers all the relationships between different representations. The modality classifier explores the corresponding modalities of features before fusion. An overview of the proposed method is given in Fig.~\ref{fig:network}. In this section, we introduce the three parts in detail.

\subsection{Diversified Attention Feature Projector}
\label{ssec:multisemanticfeature}

In the feature encoding stage, we first use the denoising operation in the input features \cite{zhan2018comprehensive}, which sets part of the input elements as zero to reduce the negative influence of noises. Then we use the weight sharing strategy to project the image and text features to the common representation space to enhance the invariant properties of the same object. To get multiple representations of an instance, we extract $k$ representations of an instance by reshaping the former features, and then fuse the generated $k$ representations with the generated attention maps by diversified attention including self-attention and co-attention, where the attention map of each instance is generated by the Eq.~\ref{eqn:atten}. The self-attention is adopted to consider the interaction of two features in the same modality, and the co-attention combines the synergistic significance of image and text so that features of different modalities can be guided to each other. The operations of self-attention and co-attention are defined as Eq.~\ref{eqn:se} and Eq.~\ref{eqn:co}, which represent the final representations of $\textbf{x}_i$.
\begin{align}
 & \textbf{a} = softmax((\textbf{W}_2tanh(\textbf{W}_1\textbf{x}))^T), \label{eqn:atten}\\
& \textup{Se}(\textbf{x}_i,\textbf{x}_j) = \textbf{x}_i (\textbf{a}_i^x + \textbf{a}_j^x), \label{eqn:se} \\
& \textup{Co}(\textbf{x}_i,\textbf{y}_j) = \textbf{x}_i (\textbf{a}_i^x + \textbf{a}_j^y), \label{eqn:co}
\end{align}
where $\textbf{x}$ and $\textbf{y} \in \mathbb{R}^{H \times k}$ represents image or text feature, $\textbf{W}_1 \in \mathbb{R}^{D \times H}$, $\textbf{W}_2 \in \mathbb{R}^{1 \times D}$; we set $D=H/2$. In Eq.~\ref{eqn:se}, if we calculate the final representation of image feature $\textbf{x}_i$ by self-attention, different image attention maps need to be fused. In Eq.~\ref{eqn:co}, if we calculate the final representation of image feature $\textbf{x}_i$ by co-attention, image attention map and text attention map need to be fused. 

In the following, the input denoised image and text features of all instances are denoted as $\textbf{U} = \{\textbf{u}_1,...,\textbf{u}_n\}$, $\textbf{C} =  \{\textbf{c}_1,...,\textbf{c}_n\}$, the abstract image and text features via the weight sharing strategy are denoted as $\textbf{V} = \{\textbf{v}_1,...,\textbf{v}_n\}$, $\textbf{T} =  \{\textbf{t}_1,...,\textbf{t}_n\}$, $\in \mathbb{R}^{L}$. The k representations of image and text via reshaping the abstract features are denoted as $\hat{\textbf{V}} = \{\hat{\textbf{v}}_1,...,\hat{\textbf{v}}_n\}$, $\hat{\textbf{T}} =  \{\hat{\textbf{t}}_1,...,\hat{\textbf{t}}_n\}$, $\in \mathbb{R}^{L/k \times k}$. The generated image and text attentions are denoted as $\textbf{A}^{\textup{v}} = \{\textbf{a}^{\textup{v}}_1,...,\textbf{a}^{\textup{v}}_n\}$, $\textbf{A}^{\textup{t}} = \{\textbf{a}^{\textup{t}}_1,...,\textbf{a}^{\textup{t}}_n\}$, $\in \mathbb{R}^{k \times 1}$. 

\subsection{Graph Pattern Loss}
\label{ssec:gdpae}

Instead of applying an expensive scheme to calculate all distances in the entire instance space, the graph pattern loss is performed within mini-batch. The first term of graph pattern loss considers the pairwise distances between different representations of the same objects  from  cross-modalities. The pairwise distance loss is defined as:
\begin{align}
L_{\textup{pdl}} = l_{\textup{p}}(\hat{\textbf{v}}_i,\hat{\textbf{t}}_i),
\label{eqn:pairdistance}
\end{align}
where $l_{\textup{p}}(\textbf{x}_i,\textbf{y}_i) =  d_{\textup{cos}}(\textup{Co}(\textbf{x}_i,\textbf{y}_i),\textup{Co}(\textbf{y}_i,\textbf{x}_i))$, $d_{\textup{cos}}$ uses the cosine distance as distance metric:
\begin{align}
d_{\textup{cos}}(\textbf{a},\textbf{b}) = \frac {1-\textbf{a} \cdot \textbf{b}}{\Vert \textbf{a} \Vert \Vert \textbf{b} \Vert } = 1 - \frac {\sum_{i=0}^{n}{a_ib_i}} {\sqrt{\sum_{i=0}^{n} {a_i^{2}}} \sqrt{\sum_{i=0}^{n}{b_i^{2}}}}.
\label{eqn:cosinedistance}
\end{align}

The pairwise distance reflects the similarity of the matched image-text representations,  which is expected to become smaller with the process of optimizing learning.

The unpaired distances represent the differences between the two representations extracted from the two objects between different modalities or in the same modality. We define all the unpaired distances of an instance as:
\begin{align}
l_{\textup{unp}}(\textbf{x}_i,\textbf{Y}) = \frac {1}{n}\sum_{j=1,j\not=i}^n (l_{\textup{p}}(\textbf{x}_i,\textbf{y}_j) - d),
\label{eqn:unpair}
\end{align}
where
$d = d_{\textup{ori}}/d_{\textup{mean}}$, $d_{\textup{ori}} = \sqrt{d_{\textup{cos}}(\textbf{u}_i,\textbf{u}_j)d_{\textup{cos}}(\textbf{c}_i,\textbf{c}_j)}$, $d_{\textup{mean}}$ represents the mean of all the $d_{\textup{ori}}$. If the $\textbf{x}$ and $\textbf{y}$ features are from the same modality, then $l_{\textup{p}}(\textbf{x}_i,\textbf{y}_j) =  d_{\textup{cos}}(\textup{Se}(\textbf{x}_i,\textbf{y}_j),\textup{Se}(\textbf{y}_j,\textbf{x}_i))$.

We take into account all the unpaired distances and limit them to be consistent with the distances between the corresponding objects in their original media spaces. We define the unpaired distance preserving loss as:
\begin{align}
L_{\textup{udp}} = l_{\textup{unp}}(\hat{\textbf{v}}_i,\hat{\textbf{T}}) + l_{\textup{unp}}(\hat{\textbf{v}}_i,\hat{\textbf{V}}) + l_{\textup{unp}}(\hat{\textbf{t}}_i,\hat{\textbf{T}}).
\label{eqn:udp}
\end{align}

By the Eq.~\ref{eqn:udp}, all the unpaired distances between different representations can be preserved. The two representations of different objects get close only if the distances between those representations are small in both the original image and text spaces, otherwise, those representations do not cluster together.

Each pair of representations involved in the three unpaired distances mentioned above come from the same objects, so all distances between them should be consistent. We keep these three unpaired distances in line with each other to further emphasize the relationship between cross-modal instances. We define the mutual distance preserving loss as:
\begin{align}
L_{\textup{mdp}} = & \vert l_{\textup{p}}(\hat{\textbf{v}}_i, \hat{\textbf{t}}_j) - l_{\textup{p}}(\hat{\textbf{v}}_i, \hat{\textbf{v}}_j) \vert + \vert l_{\textup{p}}(\hat{\textbf{v}}_i, \hat{\textbf{t}}_j) - l_{\textup{p}}(\hat{\textbf{t}}_i, \hat{\textbf{t}}_j) \vert \nonumber \\
& + \vert l_{\textup{p}}(\hat{\textbf{v}}_i, \hat{\textbf{v}}_j) - l_{\textup{p}}(\hat{\textbf{t}}_i, \hat{\textbf{t}}_j) \vert.
\label{eqn:mutualmodal}
\end{align}

Finally, we combine all the distance losses mentioned above to generate a graph form. The distance losses take into account the relationship between each representation, thereby constraining all representations, whether they come from cross-modal or single-modal and whether they belong to the same object or not. Consequently, the graph pattern loss is defined as:
\begin{align}
L_{\textup{gpl}} = L_{\textup{pdl}} + \alpha L_{\textup{udp}} + \beta L_{\textup{mdp}}
\label{eqn:joinloss},
\end{align}
where $\alpha$ and $\beta$ are the parameters to adjust the weights of the pairwise distance loss, unpaired distance preserving loss, and mutual distance preserving loss.

In the testing stage, it should be noted that only the co-attention is needed due to cross-modal retrieval.

\subsection{Modality Classifier}
\label{ssec:modalityclassifier}

In the above, we have discussed how to build the graph pattern loss. To further help the learning of cross-modal, a modality classifier is added to explicitly declare the corresponding modalities of features before fusion. We train modality classifier-D to maximize the probability of assigning the ground-truth modality label for both image and text representations from feature projector-G:
\begin{align}
L_{\textup{D}} =l_{\textup{cel}}(\textup{D}(\textbf{v}_i),\textbf{y}^1) + l_{\textup{cel}}(\textup{D}(\textbf{t} _i),\textbf{y}^0),
\label{eqn: adversarialloss}
\end{align}
where $\textbf{y}$ is the ground-truth modality label of each instance and is expressed as a one-hot vector. $\textup{D}(\ast)$ is the generated modality probability of the instance. $l_{\textup{cel}}$ represents the cross-entropy loss, $l_{\textup{cel}}= -(\textbf{y} \log \textbf{x} +(\textbf{1}-\textbf{y}) \log(\textbf{1}- \textbf{x}))$.

Given the modality classifier-D, the training of the feature projector-G minimizes the cross-entropy loss and strengthens the stability of the graph pattern loss. The loss relevant to the feature projector-G is the combination of the graph pattern loss and cross-entropy loss:
\begin{align}
L_{\textup{G}} = L_{\textup{gpl}} + \lambda (l_{\textup{cel}}(\textup{D}(\textbf{v}_{i}),\textbf{y}^{0}) + l_{\textup{cel}}(\textup{D}(\textbf{t}_{i}),\textbf{y}^{1})).
\label{eqn: lossg}
\end{align}

\section{EXPERIMENTS}
\label{sec:result}

\subsection{Dataset}
\label{sec:dataset}

We evaluate the proposed method on the four publicly available datasets: Wikipedia \cite{rasiwasia2010new}, Pascal Sentence \cite{rashtchian2010collecting}, NUS-WIDE-10K \cite{chua2009nus}, XMedia \cite{peng2015semi}. For the first three datasets, we use the 4,096d vector of the fc7 layer of the Pre-training VGG-19 \cite{simonyan2014very} for each image and 1,024d vector of the Pre-training BERT model \cite{devlin2018bert} for each text. Since the official website of the XMedia dataset does not provide the original image and text, we use the 4, 096d vector of the fc7 layer of the AlexNet \cite{krizhevsky2012imagenet} for each image and 3,000d bag-of-words(BOW) vectors for each text. The statistical results of the four datasets are summarised in Table~\ref{tab:dataset}. 

\begin{table}[ht]
\footnotesize
\setlength{\abovecaptionskip}{0.cm}
\setlength{\belowcaptionskip}{-0.cm}
\caption{General statistics of the four datasets used in our experiments, the second column stands for the number of training/test image-text instances.}
\label{tab:dataset}
\centering
\begin{tabular}{l|llll}
\hline
Dataset& Instance& Labels& $d_{i}$& $d_{i}$\\
\hline
Wikipedia& 2,173/462& 10& 4096& 1024\\
Pascal Sentence& 800/100& 20& 4096& 1024\\
NUS-WIDE-10k& 8,000/1,000& 10& 4096& 1024\\
XMedia& 4,000/700& 20& 4096& 3000\\
\hline
\end{tabular}
\end{table}

\subsection{Implementation Details}
\label{sec:ImplementationDetails}

The GPLDAN is implemented using the Pytorch package \cite{paszke2017automatic}. We use Adam optimizer for training the feature projector with a learning rate of $10^{-4}$ and weight decay of $10^{-4}$, and RMSprop optimizer for training the modality classifier with a learning rate of $5 \times10^{-5}$. The parameters of loss: the values of $\lambda$ is fixed at 0.01, $\alpha$ and $\beta$ are tuned by using a grid search. The best-reported results of GPLDAN are obtained for the optimal values of $\alpha$ and $\beta$ per dataset. The parameter $k$ is fixed at 4 in the diversified attention feature projector.

\subsection{Comparison with Existing Methods}
\label{sec:Comparison}

In this subsection, to evaluate the effectiveness of the proposed method, we compare the GPLDAN with eight cross-modal retrieval methods. The MAP@50 scores for all the cross-modal retrieval tasks are shown in Table~\ref{tab:ablation}. For each measurement, the top three methods are highlighted in boldface.

The GPLDAN nearly outperforms all the unsupervised methods. Only for the Text2Image task in the XMedia, GPLDAN is 0.5\% lower than the best method, ranking second. The GPLDAN still performs competitively when compared with the semi-supervised and the supervised methods. It can be observed that the supervised methods generally outperform unsupervised methods in all retrieval tasks. However, the GPLDAN ranks in the top three in most retrieval tasks. In particular, in both retrieval tasks for Pascal Sentence and Txt2Img for NUS-WIDE-10K, the GPLDAN outperforms all the other methods. The performance of the GPLDAN is relatively robust. Other methods work well for some datasets but fail to provide satisfactory results for the rest. The GPLDAN can be flexible to be applied for diverse image features extracted by different pre-trained CNN and diverse text features extracted by Bert or BOW.

\begin{table*}[ht]
\footnotesize
\setlength{\abovecaptionskip}{0.cm}
\setlength{\belowcaptionskip}{-0.cm}
\caption{Performance comparison of cross-modal retrieval methods on four datasets.}
\label{tab:ablation}
\centering
\begin{tabular}{c|c|ccc|c|ccccc}
\hline 
\multirow{2}{*}{Datasets}&\multirow{2}{*}{Tasks}
&\multicolumn{3}{c|}{Supervised}
&\multicolumn{1}{c|}{Semi-supervised }
&\multicolumn{5}{c}{Unsupervised}\\
&&{CMCP} & {ACMR} & {DSCMR} &{JRL} & {CCA}& {DCCA}& {DCCAE}& {CDPAE}& {\textbf{GPLDAN}}\\
\hline
\multirow{3}{*}{Wikipedia}
& {Img2Txt} & \textbf{0.499}& 0.443& \textbf{0.493}& 0.472& 0.296& 0.388& 0.361& 0.470& \textbf{0.482}\\
& {Txt2Img} & \textbf{0.593}& 0.499& \textbf{0.632}& \textbf{0.560}& 0.346& 0.426&	0.425& 0.475& 0.543\\ 
& {Avg.} & \textbf{0.546}& 0.471& \textbf{0.563}& \textbf{0.516}& 0.321& 0.407& 0.393& 0.473&  0.513\\
\hline
\multirow{3}{*}{Pascal Sentence}
& {Img2Txt} & 0.378& \textbf{0.407}& 0.392& 0.341&  0.279& 0.366& 0.405& \textbf{0.413}& \textbf{0.476}\\
& {Txt2Img} & 0.355& 0.410& \textbf{0.415}& 0.359& 0.256& 0.354& \textbf{0.413}& 0.353&  \textbf{0.462}\\ 
& {Avg.} & 0.366& \textbf{0.408}& 0.404& 0.350& 0.268& 0.360& \textbf{0.409}& 0.383& \textbf{0.469}\\
\hline
\multirow{3}{*}{NUS-WIDE-10K}
& {Img2Txt} & 0.378& 0.306& \textbf{0.408}& \textbf{0.400}& 0.258& 0.346& 0.329& 0.383&  \textbf{0.389}\\
& {Txt2Img} & 0.334& 0.308& \textbf{0.373}& \textbf{0.376}& 0.250& 0.337& 0.337& 0.368& \textbf{0.381}\\ 
& {Avg.} & 0.356& 0.307& \textbf{0.390}& \textbf{0.388}& 0.254& 0.342& 0.333& 0.376& \textbf{0.385}\\
\hline
\multirow{3}{*}{XMedia}
& {Img2Txt} & 0.831& 0.895& \textbf{0.911}& 0.899& 0.122& 0.762& 0.772& \textbf{0.901}& \textbf{0.904}\\
& {Txt2Img} & 0.765& 0.931& \textbf{0.935}& 0.934& 0.120& 0.801& 0.773&  \textbf{0.947}& \textbf{0.942}\\ 
& {Avg.} & 0.798& 0.913& \textbf{0.923}& 0.917& 0.121& 0.773& 0.781& \textbf{0.924}& \textbf{0.923}\\
\hline
\end{tabular}
\end{table*}

\subsection{Ablation Analysis of GPLDAN}
\label{sec:gsdpa}

\subsubsection{Model Parameters Analysis}
\label{sssec:mpa}

\begin{figure}[h]
\setlength{\abovecaptionskip}{0.cm}
\setlength{\belowcaptionskip}{-0.cm}
     \begin{minipage}{0.5\linewidth}
            \centering 
	    \includegraphics[width=1.6in]{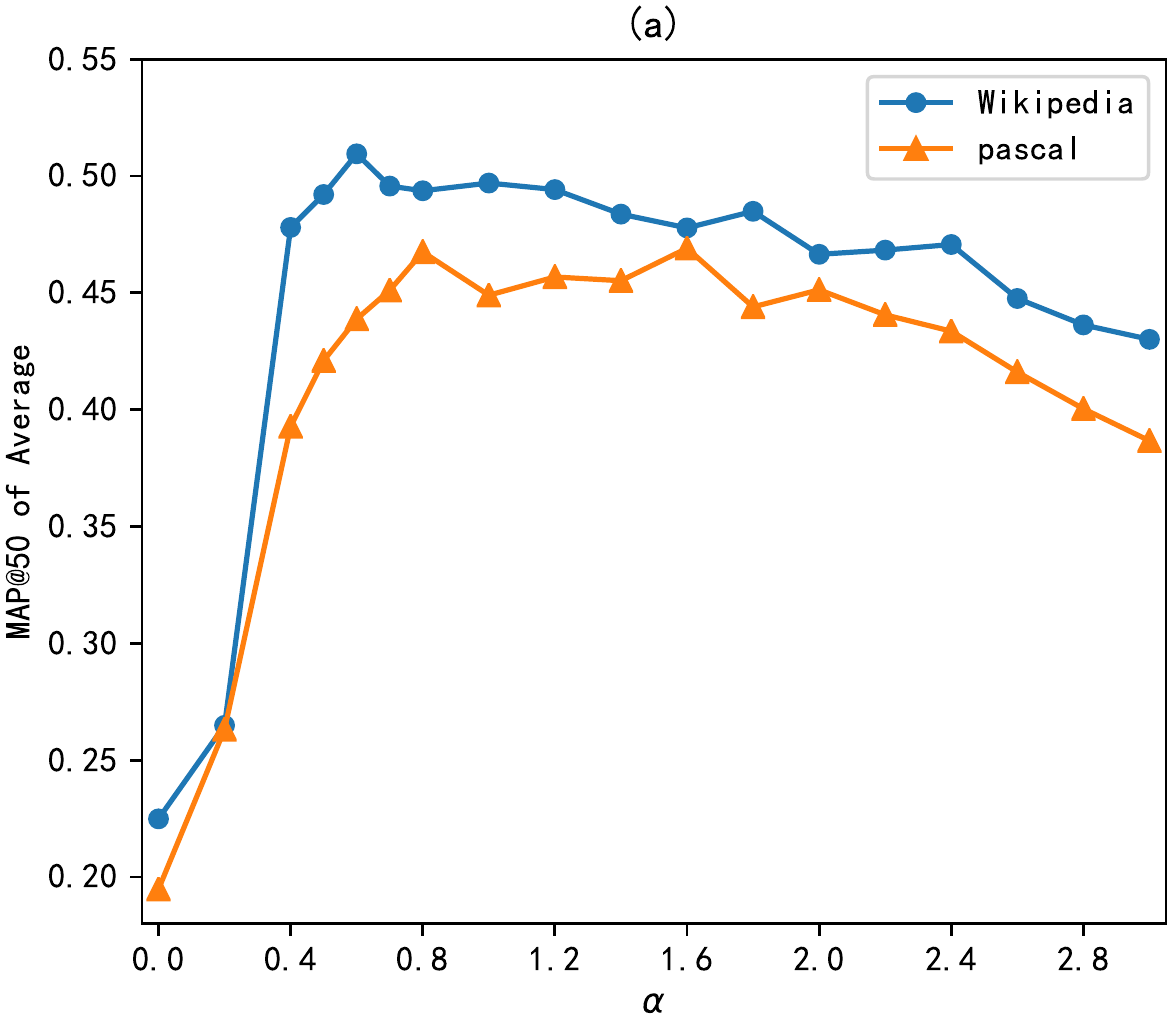} 
     \end{minipage}
     \hfill 
     \begin{minipage}{0.5\linewidth}
           \centering 
     	   \includegraphics[width=1.6in]{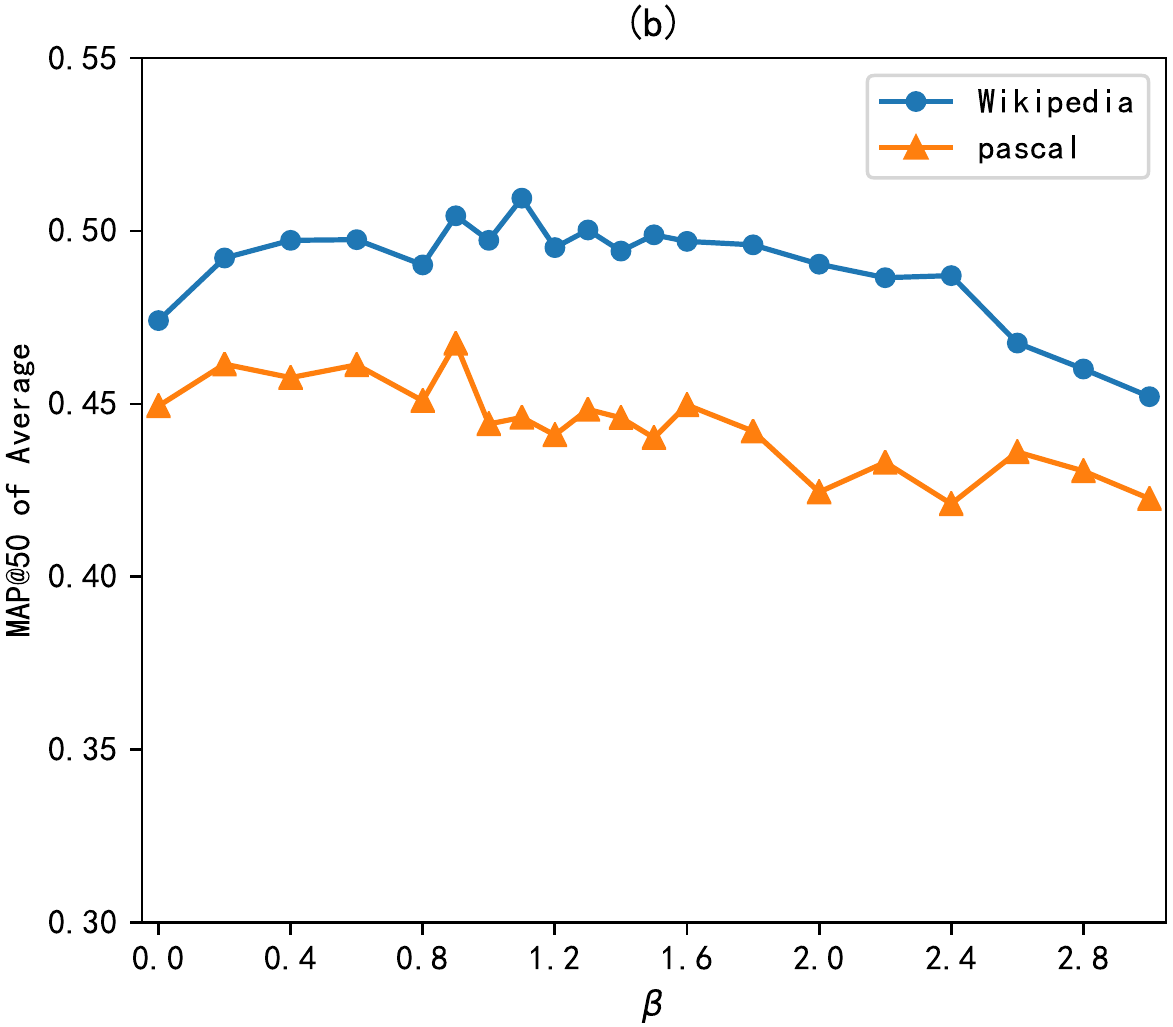}
     \end{minipage}
\caption{The performance of the GPLDAN with different values: (a) $\alpha$ and (b) $\beta$.}
\label{fig:aabb}
\end{figure}

In previous experiments, we empirically set the values of the $\alpha$ and $\beta$. To explore the impacts of different values, Wikipedia and Pascal Sentence datasets are used as test cases. The evaluation is conducted by changing one of the observed parameters while fixing another. Figure.~\ref{fig:aabb} shows the average performance of the GPLDAN with different values. From Figure.~\ref{fig:aabb}, we can draw the following conclusions:

1) $\alpha$ determines the significance of the unpaired distance preserving loss. If $\alpha = 0$, graph pattern loss only considers the pairwise distance between representations of the same objects from different modalities and the mutual distance preserving loss. However, only providing mutual distance does not give the correct guidance of the unpaired distance. From Figure.~\ref{fig:aabb}-(a), we can observe that both higher or lower values of $\alpha$ result in poor performance. Based on the above analysis, the unpaired distance preserving loss plays an important role in cross-modal retrieval.

2) $\beta$ represents the significance of the mutual distance preserving loss. From Figure.~\ref{fig:aabb}-(b), we can find that the mutual distance preserving loss contributes slightly to the final retrieval performance. However, the final result can be improved by giving the appropriate parameters of the mutual distance preserving loss.

\subsubsection{Components Analysis}
\label{sssec:ca}

To explore the importance of different components, we test the performances by gradually adding components. The results are presented in Table~\ref{tab:casis}. The baseline system only includes pairwise distance loss. From the results, we can find that the performance has been increased correspondingly with the addition of the unpaired distance preserving loss, mutual distance preserving loss, diversified attention, and modality classifier. When we use the unpaired distance preserving loss on the baseline system, we find that the performance is improved a lot i.e., 24.8\%. When the mutual distance preserving loss is added to the baseline system, the result is rose by 3.1\%. If the mutual distance preserving loss and the unpaired distance preserving loss are simultaneously added to the baseline system, the result can be further improved, which shows that the mutual distance preserving loss can further emphasize the relationship of cross-modal instances, but these three losses need to work together and constrain each other. If the modality classifier is applied, the result is increased from 0.487 to 0.496. Furthermore, when the diversified attention is used, the result is improved from 0.496 to 0.513, which proves that diversified attention can help the network better consider the interaction between different representations and generate useful features.

\begin{table}[ht]
\footnotesize
\setlength{\abovecaptionskip}{0.cm}
\setlength{\belowcaptionskip}{-0.cm}
\caption{The performance of our method on Wikipedia dataset under different constraints. MC denotes modality classifier, DA denotes diversified attention.}
\label{tab:casis}
\centering
\begin{tabular}{l|lll}
\hline
Method& Img2Txt& Txt2Img& Avg.\\
\hline
Baseline& 0.189& 0.215& 0.202\\
$+ L_{\textup{mdp}}$& 0.249& 0.217& 0.233\\
$+ L_{\textup{udp}}$& 0.456& 0.506& 0.481\\
$+ L_{\textup{mdp}}, L_{\textup{udp}}$& 0.462& 0.512& 0.487\\
$+ L_{\textup{mdp}}, L_{\textup{udp}}$, MC& 0.476& 0.516& 0.496\\
$+ L_{\textup{mdp}}, L_{\textup{udp}}$, MC, DA (GPLDAN)& \textbf{0.482}& \textbf{0.543}& \textbf{0.513}\\
\hline
\end{tabular}
\end{table}

\section{CONCLUSIONS}
\label{sec:conclusion}

In this work, we propose a Graph Pattern Loss based Diversified Attention Network (GPLDAN) for unsupervised cross-modal retrieval. Our diversified attention feature projector can extract multiple representations of an instance by considering the interaction between different representations. In the graph pattern loss, all the correlations among different representations are explored and all possible distances between different representations are considered. The modality classifier is used to explicitly declare the corresponding modalities of features before fusion. Our approach nearly outperforms all the unsupervised retrieval methods and performs competitively when compared with the semi-supervised and the supervised methods. Besides, our approach has a good generation ability and can be applied to different modal representations.
\bibliographystyle{IEEEbib}
\bibliography{strings,refs}

\begin{thebibliography}{10}

\bibitem{wang2017adversarial}
Bokun Wang, Yang Yang, Xing Xu, Alan Hanjalic, and Heng~Tao Shen,
\newblock ``Adversarial cross-modal retrieval,''
\newblock in {\em Proceedings of the 25th ACM international conference on
  Multimedia}. ACM, 2017, pp. 154--162.

\bibitem{wang2016comprehensive}
Kaiye Wang, Qiyue Yin, Wei Wang, Shu Wu, and Liang Wang,
\newblock ``A comprehensive survey on cross-modal retrieval,''
\newblock {\em arXiv preprint arXiv:1607.06215}, 2016.

\bibitem{peng2017ccl}
Yuxin Peng, Jinwei Qi, Xin Huang, and Yuxin Yuan,
\newblock ``Ccl: Cross-modal correlation learning with multigrained fusion by
  hierarchical network,''
\newblock {\em IEEE Transactions on Multimedia}, vol. 20, no. 2, pp. 405--420,
  2017.

\bibitem{rasiwasia2010new}
Nikhil Rasiwasia, Jose Costa~Pereira, Emanuele Coviello, Gabriel Doyle, Gert~RG
  Lanckriet, Roger Levy, and Nuno Vasconcelos,
\newblock ``A new approach to cross-modal multimedia retrieval,''
\newblock in {\em Proceedings of the 18th ACM international conference on
  Multimedia}. ACM, 2010, pp. 251--260.

\bibitem{wang2015joint}
Kaiye Wang, Ran He, Liang Wang, Wei Wang, and Tieniu Tan,
\newblock ``Joint feature selection and subspace learning for cross-modal
  retrieval,''
\newblock {\em IEEE transactions on pattern analysis and machine intelligence},
  vol. 38, no. 10, pp. 2010--2023, 2015.

\bibitem{wang2016learning}
Liwei Wang, Yin Li, and Svetlana Lazebnik,
\newblock ``Learning deep structure-preserving image-text embeddings,''
\newblock in {\em Proceedings of the IEEE conference on computer vision and
  pattern recognition}, 2016, pp. 5005--5013.

\bibitem{zhai2012cross}
Xiaohua Zhai, Yuxin Peng, and Jianguo Xiao,
\newblock ``Cross-modality correlation propagation for cross-media retrieval,''
\newblock in {\em 2012 IEEE International Conference on Acoustics, Speech and
  Signal Processing (ICASSP)}. IEEE, 2012, pp. 2337--2340.

\bibitem{Zhen_2019_CVPR}
Liangli Zhen, Peng Hu, Xu~Wang, and Dezhong Peng,
\newblock ``Deep supervised cross-modal retrieval,''
\newblock in {\em The IEEE Conference on Computer Vision and Pattern
  Recognition (CVPR)}, June 2019.

\bibitem{peng2015semi}
Yuxin Peng, Xiaohua Zhai, Yunzhen Zhao, and Xin Huang,
\newblock ``Semi-supervised cross-media feature learning with unified patch
  graph regularization,''
\newblock {\em IEEE Transactions on Circuits and Systems for Video Technology},
  vol. 26, no. 3, pp. 583--596, 2015.

\bibitem{zhai2013learning}
Xiaohua Zhai, Yuxin Peng, and Jianguo Xiao,
\newblock ``Learning cross-media joint representation with sparse and
  semisupervised regularization,''
\newblock {\em IEEE Transactions on Circuits and Systems for Video Technology},
  vol. 24, no. 6, pp. 965--978, 2013.

\bibitem{hotelling1992relations}
Harold Hotelling,
\newblock ``Relations between two sets of variates,''
\newblock in {\em Breakthroughs in statistics}, pp. 162--190. Springer, 1992.

\bibitem{andrew2013deep}
Galen Andrew, Raman Arora, Jeff Bilmes, and Karen Livescu,
\newblock ``Deep canonical correlation analysis,''
\newblock in {\em International conference on machine learning}, 2013, pp.
  1247--1255.

\bibitem{feng2014cross}
Fangxiang Feng, Xiaojie Wang, and Ruifan Li,
\newblock ``Cross-modal retrieval with correspondence autoencoder,''
\newblock in {\em Proceedings of the 22nd ACM international conference on
  Multimedia}. ACM, 2014, pp. 7--16.

\bibitem{he2017unsupervised}
Li~He, Xing Xu, Huimin Lu, Yang Yang, Fumin Shen, and Heng~Tao Shen,
\newblock ``Unsupervised cross-modal retrieval through adversarial learning,''
\newblock in {\em 2017 IEEE International Conference on Multimedia and Expo
  (ICME)}. IEEE, 2017, pp. 1153--1158.

\bibitem{wang2015deep}
Weiran Wang, Raman Arora, Karen Livescu, and Jeff Bilmes,
\newblock ``On deep multi-view representation learning,''
\newblock in {\em International Conference on Machine Learning}, 2015, pp.
  1083--1092.

\bibitem{wang2014effective}
Wei Wang, Beng~Chin Ooi, Xiaoyan Yang, Dongxiang Zhang, and Yueting Zhuang,
\newblock ``Effective multi-modal retrieval based on stacked auto-encoders,''
\newblock {\em Proceedings of the VLDB Endowment}, vol. 7, no. 8, pp. 649--660,
  2014.

\bibitem{yan2015deep}
Fei Yan and Krystian Mikolajczyk,
\newblock ``Deep correlation for matching images and text,''
\newblock in {\em Proceedings of the IEEE conference on computer vision and
  pattern recognition}, 2015, pp. 3441--3450.

\bibitem{zhan2018comprehensive}
Yibing Zhan, Jun Yu, Zhou Yu, Rong Zhang, Dacheng Tao, and Qi~Tian,
\newblock ``Comprehensive distance-preserving autoencoders for cross-modal
  retrieval,''
\newblock in {\em 2018 ACM Multimedia Conference on Multimedia Conference}.
  ACM, 2018, pp. 1137--1145.

\bibitem{goodfellow2014generative}
Ian Goodfellow, Jean Pouget-Abadie, Mehdi Mirza, Bing Xu, David Warde-Farley,
  Sherjil Ozair, Aaron Courville, and Yoshua Bengio,
\newblock ``Generative adversarial nets,''
\newblock in {\em Advances in neural information processing systems}, 2014, pp.
  2672--2680.

\bibitem{rashtchian2010collecting}
Cyrus Rashtchian, Peter Young, Micah Hodosh, and Julia Hockenmaier,
\newblock ``Collecting image annotations using amazon's mechanical turk,''
\newblock in {\em Proceedings of the NAACL HLT 2010 Workshop on Creating Speech
  and Language Data with Amazon's Mechanical Turk}. Association for
  Computational Linguistics, 2010, pp. 139--147.

\bibitem{chua2009nus}
Tat-Seng Chua, Jinhui Tang, Richang Hong, Haojie Li, Zhiping Luo, and Yantao
  Zheng,
\newblock ``Nus-wide: a real-world web image database from national university
  of singapore,''
\newblock in {\em Proceedings of the ACM international conference on image and
  video retrieval}. ACM, 2009, p.~48.

\bibitem{simonyan2014very}
Karen Simonyan and Andrew Zisserman,
\newblock ``Very deep convolutional networks for large-scale image
  recognition,''
\newblock {\em arXiv preprint arXiv:1409.1556}, 2014.

\bibitem{devlin2018bert}
Jacob Devlin, Ming-Wei Chang, Kenton Lee, and Kristina Toutanova,
\newblock ``Bert: Pre-training of deep bidirectional transformers for language
  understanding,''
\newblock {\em arXiv preprint arXiv:1810.04805}, 2018.

\bibitem{krizhevsky2012imagenet}
Alex Krizhevsky, Ilya Sutskever, and Geoffrey~E Hinton,
\newblock ``Imagenet classification with deep convolutional neural networks,''
\newblock in {\em Advances in neural information processing systems}, 2012, pp.
  1097--1105.

\bibitem{paszke2017automatic}
Adam Paszke, Sam Gross, Soumith Chintala, Gregory Chanan, Edward Yang, Zachary
  DeVito, Zeming Lin, Alban Desmaison, Luca Antiga, and Adam Lerer,
\newblock ``Automatic differentiation in pytorch,''
\newblock 2017.

\end{thebibliography}

\end{document}